%% file: main.tex
\documentclass[runningheads]{llncs}

\PassOptionsToPackage{table}{xcolor}
\usepackage[year=2026]{eccv}

\usepackage{eccvabbrv}

\usepackage{graphicx}
\usepackage[accsupp]{axessibility}  

\usepackage{hyperref}
\usepackage{orcidlink}

\usepackage{booktabs}      
\usepackage{multirow}      
\usepackage{siunitx}       
\usepackage{xcolor} 
\usepackage{adjustbox}     
\newcommand{\mch}[1]{\multicolumn{1}{c}{#1}}
\newcommand{\best}[1]{\textbf{#1}}
\newcommand{\secbest}[1]{\underline{#1}}

\usepackage{tikz}
\usepackage{makecell}
\newcommand{\lpipscell}[1]{\multicolumn{1}{>{\columncolor{black!8}}S[table-format=2.4]}{#1}}

\newcommand{\synthcell}[1]{\multicolumn{1}{>{\columncolor{eccvblue!10}}S[table-format=2.4]}{#1}}
\newcommand{\synthcellDark}[1]{\multicolumn{1}{>{\columncolor{eccvblue!20}}S[table-format=2.4]}{#1}}
\newcommand{\fullcell}[1]{\multicolumn{1}{>{\columncolor{eccvblue!10}}S[table-format=2.4]}{#1}}

\newcommand{\lpipscellbluebox}[1]{%
  \multicolumn{1}{>{\columncolor{black!8}}c}{%
    \setlength{\fboxrule}{1.2pt}
    \setlength{\fboxsep}{1.5pt}
    \fcolorbox{eccvblue}{black!8}{#1}
  }%
}

\usepackage{pifont}
\newcommand{\cmark}{\ding{51}}
\newcommand{\xmark}{\ding{55}}
\usepackage{microtype}
\usepackage{placeins} 
\begin{document}

\input{paper}
\bibliographystyle{splncs04}
\bibliography{main}
\clearpage
\appendix
\setcounter{figure}{0}
\renewcommand{\thefigure}{A\arabic{figure}}
\renewcommand{\theHfigure}{appendix.\arabic{figure}}
\input{appendix}
\end{document}

%% file: paper.tex
\title{Fill2SR: Repurposing Inpainting Diffusion Transformers for Real-World Super-Resolution}

\titlerunning{Fill2SR}

\author{Xingfu Yi\inst{1}\orcidlink{0000-0002-5792-6280} \and
Xiaoxue Yu\inst{2}\orcidlink{0009-0008-1098-7589}}

\authorrunning{X.~Yi and X.~Yu}

\institute{Independent Researcher, Hangzhou, China\\
\email{yixingfu.research@gmail.com} \and
Zhejiang University, Hangzhou, China}

\maketitle

\begin{figure}[h]
    \centering
    \includegraphics[width=\textwidth]{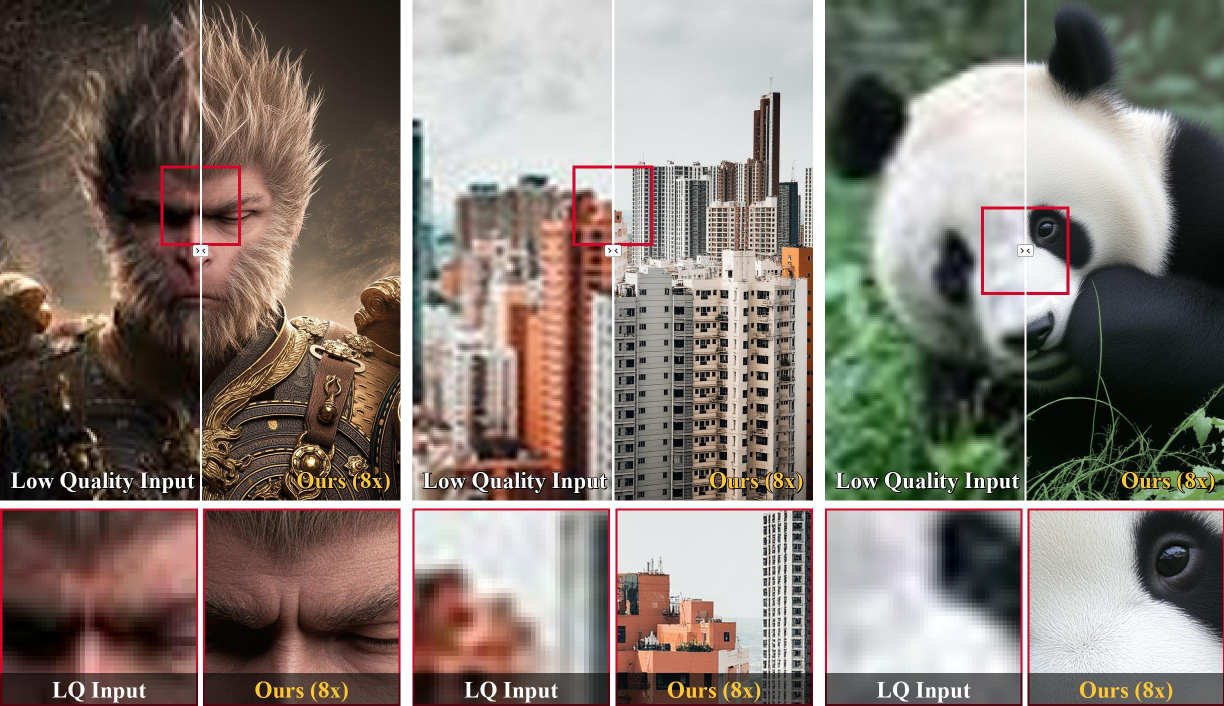}
    \caption{
    \textbf{Visual highlights on \textit{RealLQ250} at $8\times$ (real-world SR).}
    Fill2SR produces artifact-suppressed, visually coherent high-resolution results across diverse content: non-rigid textures (left, Wukong), rigid urban structures (middle, Cityscapes), and fine biological details (right, Panda).
    \emph{Bottom:} zoom-ins from the red boxes.
    }
    \label{fig:teaser}
\end{figure}

\begin{abstract}
Recent real-world image super-resolution (SR) methods often adapt text-to-image (T2I) backbones with ControlNet-style branches or spatial conditioning tokens, which increases memory and computes with resolution and often constrains training to a fixed scale. We propose \textbf{Fill2SR}, which repurposes a masked-inpainting Diffusion Transformer for SR without extra spatial branches. Our Inpainting-Interface Evidence Adapter (\textbf{IIEA}) writes the low-quality (LQ) observation into the native masked-image slot under a full-image mask, turning inpainting into a reverse-degradation conditional rectified flow trained with LoRA-only tuning. We further introduce \textbf{RCDT}, an offline pipeline that distills degradation descriptors from unpaired real images and transfers them onto clean targets using frozen open-source models. Fill2SR supports mixed-resolution training up to QHD and yields stable performance across $512/1024/2048$ outputs. On synthetic benchmarks, our base model with IIEA achieves the best LPIPS on DIV2K and LSDIR; adding RCDT trades a small LPIPS drop for consistently stronger no-reference quality on RealLQ250 and RealPhoto60. Fill2SR remains memory-predictable, running $1536^2$ inference on a single 32GB GPU and extending to multi-megapixel outputs via tiled restoration.
\keywords{Real-World Super-Resolution \and Image Restoration \and Inpainting Diffusion Transformers \and Rectified Flow Matching \and Adapter Tuning}
\end{abstract}

\section{Introduction}
Image super-resolution (SR) and restoration recover high-quality (HQ) images from low-quality (LQ) observations corrupted by blur, noise, compression, and complex real-world degradations. Despite strong discriminative restorers such as SwinIR~\cite{SwinIR}, Restormer~\cite{zamir2022restormer}, and Uformer~\cite{wang2022uformer}, distortion-driven training often follows the perception-distortion trade-off and produces over-smoothed details under severe degradations~\cite{Blau_2018_CVPR}. To train these models, previous approaches predominantly rely on artificially synthesized LQ-HQ data pairs. While synthetic degradation pipelines (e.g., BSRGAN~\cite{zhang2021bsrgan}, Real-ESRGAN~\cite{wang2021realesrgan}) offer scalable supervision for this paradigm, they fundamentally struggle to cover the long tail of complex corruptions in the wild.

Generative diffusion models~\cite{DDPM,iDDPM,beatGANs} provide strong image priors and have become a leading approach for high-fidelity restoration. However, synthesis-oriented text-to-image (T2I) backbones are not optimized for pixel-aligned control, making them prone to geometric drift or spurious artifacts under severe degradations. Consequently, recent approaches typically train ControlNet-style branches~\cite{SeeSR, SUPIR,FaithDiff} or introduce additional spatial conditioning tokens~\cite{DreamClear, dit4sr, FluxIR}. While these workarounds offer improvements, T2I-centric restoration pipelines face two structural bottlenecks. First, the memory/compute overhead induced by extra branches or tokens grows superlinearly with resolution (often quadratically in the number of tokens), strictly limiting the training resolution. Second, due to memory limits~\cite{SUPIR,FaithDiff,dit4sr}, the control module is typically trained at a specific resolution (e.g., $1024^2$). It matches the backbone's prior at this specific resolution but fails to generalize. As a result, these methods 
achieve good restoration quality 
at their training resolution but suffer from severe hallucinations at different scales (see Tab.~\ref{tab:real_sr_scales}, Fig.~\ref{fig:qual_realworld}, and Supplementary Material for per-degradation quantitative comparisons).

In this work, we advocate building real-world restoration on an interface that is natively pixel-aligned. We observe that modern mask-conditioned inpainting Diffusion Transformers (DiTs)~\cite{flux1_fill_dev} are trained to preserve known pixels while completing unknown regions. This induces a strong bias toward spatially consistent completion, making inpainting backbones a natural match for SR/restoration, 
since these tasks require respecting the LQ observation as strict pixel-level evidence rather than weak high-level hints. We therefore present \textbf{Fill2SR}, a low-overhead adaptation framework that repurposes a pre-trained inpainting DiT for SR and real-world restoration. Our key component is IIEA (Inpainting-Interface Evidence Adapter), which repurposes the native masked-image slot as a dense, pixel-aligned evidence channel under a full-image mask, avoiding ControlNet-style branches or resolution-growing spatial conditioning token streams. We adapt the inpainting prior to the endpoint sampling of a reverse-degradation conditional rectified flow with LoRA-only~\cite{Lora} tuning. This flow starts from pure noise and ends at the HQ endpoint of an implicit degradation model, formulated via Rectified Flow Matching (RFM)~\cite{liu2022flow,lipman2022flow}. Because real-world degradations are diverse and paired LQ-HQ supervision is scarce, synthetic degradations alone may not cover the long tail and can introduce domain bias. To address this cost-effectively, we introduce \textbf{RCDT} (Reference-Conditioned Degradation Transfer), an offline paired-synthesis pipeline that distills degradation descriptors from an unpaired real-LQ pool and transfers them onto clean HQ targets using frozen open-source models~\cite{Qwen3-VL,wu2025qwenimagetechnicalreport}. To further bound test-time conditioning cost, we replace heavyweight vision-language prompting~\cite{LLAVA} with a lightweight image embedder (Redux)~\cite{flux1_redux_dev}, stabilizing semantic control with fixed-length visual-semantic tokens rather than computationally expensive test-time image prompting.

Extensive experiments show that Fill2SR achieves strong perceptual quality on synthetic benchmarks: our base model with IIEA and trained with synthetic only data~\cite{wang2021realesrgan,FaithDiff} attains the best LPIPS on DIV2K~\cite{DIV2K} and LSDIR~\cite{LSDIR}. Incorporating RCDT-generated pairs in the final 20\% of training improves no-reference perceptual metrics~\cite{MUSIQ,MANIQA,CLIPIQA} on RealLQ250~\cite{DreamClear} and RealPhoto60~\cite{SUPIR,FaithDiff}, trading a small drop in synthetic LPIPS. In addition, the bounded conditioning cost of IIEA enables mixed-resolution, mixed-aspect-ratio training up to native QHD ($2560\times1440$), while still fitting within a single 32GB GPU at $1536^2$ for inference.
\paragraph{Contributions.}
Our main contributions are:
\begin{itemize}
  \item \textbf{Structural shift from T2I with control to Inpainting with IIEA.} We repurpose mask-conditioned inpainting DiTs for SR/restoration by leveraging their natively pixel-aligned inpainting interface. Specifically, IIEA (Inpainting-Interface Evidence Adapter) repurposes the masked-image slot as a dense evidence channel under a full-image mask, avoiding ControlNet-style branches or resolution-growing spatial token streams while enabling stable mixed-resolution training across scales.
  \item \textbf{Reverse-degradation flow for domain shift.} We repurpose the backbone's native Rectified Flow Matching (RFM) objective. While originally trained for local region completion conditioned on preserved surroundings, we shift this formulation to a full-image reverse-degradation flow. This flow models the trajectory from pure noise to the clean HQ endpoint of an implicit degradation model.
  \item \textbf{Paired reverse-degradation supervision.} To efficiently learn this reverse-degradation flow under realistic degradations, we introduce RCDT, an offline data generation pipeline that distills real-world degradation statistics from unpaired LQ images and creates ``real-world'' LQ-HQ training pairs from clean HQ images at a low cost.
\end{itemize}

\section{Related Work}

\paragraph{Discriminative and diffusion-based restoration.}
Classical methods map LQ to HQ images using CNNs or Transformers~\cite{SwinIR,zamir2022restormer,wang2022uformer}, often relying on synthetic degradations~\cite{zhang2021bsrgan,wang2021realesrgan,liang2022dasr}. To overcome over-smoothing in deterministic models, diffusion priors are widely adopted for photo-realistic generation, evolving from iterative refinement~\cite{SR3} to highly efficient variants~\cite{yue2023resshift,wang2024sinsr,wu2024one}.

\paragraph{Adapting T2I priors for restoration.}
To tackle real-world degradations, recent methods repurpose large T2I backbones, whether UNet-based~\cite{SD,SDXL,StableSR,DiffBIR,SUPIR} or DiT-based~\cite{DreamClear,dit4sr}. To enforce fidelity, they typically introduce extra conditioning stacks (\eg, ControlNet~\cite{ControlNet}) or inject LQ guidance directly into the attention mechanism. However, these structural modifications inevitably inflate system complexity, introduce additional \emph{spatial} token streams or control branches, and often remain brittle under severe degradations. 

\paragraph{Inpainting models as a restoration foundation.}
Departing from complex control modules, diffusion inpainting naturally provides a pixel-aligned \textit{(mask, context)} interface. While explored in UNet models~\cite{lugmayr2022repaint,saharia2021palette,xie2022smartbrush} and early DiTs~\cite{zhuang2023powerpaint}, repurposing massive state-of-the-art inpainting DiTs~\cite{flux1_fill_dev} for blind restoration remains underexplored. Our work capitalizes on this by repurposing the masked-image slot as an evidence channel. Combined with rectified-flow learning~\cite{chen2018neural,liu2022flow,lipman2022flow} and the conditioning design of fixed length tokens (IIEA and optional Redux~\cite{flux1_redux_dev}), we achieve memory-predictable, high-resolution restoration that bypasses the token-length inflation plaguing attention-modified DiTs.

\section{Method}
\label{sec:method}

\begin{figure}[t]
    \centering
    \includegraphics[width=1\textwidth]{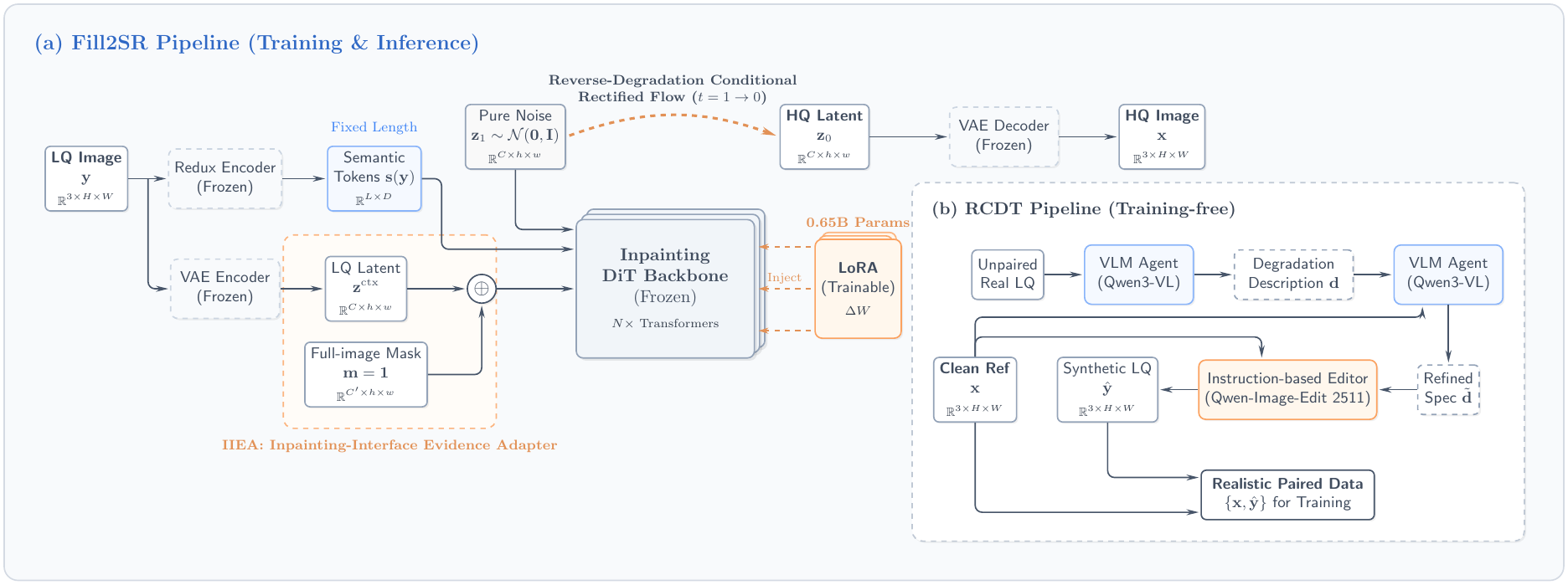}
    
    \caption{\textbf{Overview of the Fill2SR architecture.} 
    (a) The \textbf{Main Restoration Pipeline} utilizes the proposed IIEA to inject VAE-encoded LQ evidence into the native FLUX-Fill inpainting interface via a full-image mask ($\mathbf{m}=\mathbf{1}$). 
    Sampling follows a conditional rectified flow trajectory from pure noise $\mathbf{z}_1$ to the restored HQ latent $\mathbf{z}_0$.
    (b) The \textbf{Offline RCDT Pipeline} leverages multi-modal agents (e.g., Qwen3-VL) to perceive real-world degradations and generate refined specifications, driving an instruction-based editor to synthesize realistic training pairs $\{\mathbf{x}, \hat{\mathbf{y}}\}$ with zero test-time overhead.}
    \label{fig:method_overview}
\end{figure}

\paragraph{Problem setup.}
We formulate the problem and our method in the VAE ($\varepsilon$)~\cite{VAE,SD} latent space for simplicity.
Given an LQ image $Y$ in normalized [-1, 1] RGB space and its latent $y = \varepsilon(Y) \in\mathbb{R}^{c\times \frac{h}{s}\times \frac{w}{s}}$,
our goal is to get the HQ image latent $\mathbf{x} \in\mathbb{R}^{c\times h\times w }$ after an SR scale factor of $s$. 
We first upscale the LQ input so that $y \in\mathbb{R}^{c\times h\times w }$.
We consider there exists a implicit degradation model $\mathcal{T}$ that $y$ is generated by $\mathbf{y}=\mathcal{T}(\mathbf{x})$. 
Since we only have $y$ as input and our goal is to get $\mathbf{x}$, our problem is to solve the function of the reverse degradation model so that
\begin{equation}
\mathbf{x} = \mathcal{T}^{-1} (y)
\end{equation}

\subsection{Reverse-Degradation Conditional Rectified Flow}
\label{sec:reverse_flow}

Since directly modeling the real-world degradation operator $\mathcal{T}$ is intractable, we utilize Rectified Flow Matching (RFM)~\cite{liu2022flow,lipman2022flow} to learn an implicit reverse-degradation mapping. Concretely, we build a conditional rectified flow that transports pure Gaussian noise to the HQ latent $\mathbf{x}$, conditioned on the degraded observation $\mathbf{y}=\mathcal{T}(\mathbf{x})$. Following RFM, we sample an interpolation time $t\sim\mathrm{Unif}(0,1)$ and form a linear path between $\mathbf{x}$ and $\mathbf{z}_1\sim\mathcal{N}(\mathbf{0},\mathbf{I})$:
\begin{equation}
\begin{aligned}
\mathbf{z}_t &= (1-t)\mathbf{x} + t\mathbf{z}_1,\qquad \mathbf{z}_1\sim\mathcal{N}(\mathbf{0},\mathbf{I}),\quad t\sim\mathrm{Unif}(0,1),\\
\frac{d\mathbf{z}_t}{dt} &= \mathbf{z}_1 - \mathbf{x}
= \mathbf{v}_{\theta}\!\left(\mathbf{z}_t,\mathcal{T}(\mathbf{x}),t\right).
\end{aligned}
\label{eq:ode_t}
\end{equation}

For synthetic degradation, we explicitly formulate $\mathcal{T}(\mathbf{x})$ as 
\begin{equation}
\mathcal{T}(\mathbf{x})
= \mathrm{JPEG}_q\!\big(( \mathbf{x} * \kappa_{\sigma_b} ) \downarrow_s \big) + \eta, 
\quad \eta \sim \mathcal{N}(0,\sigma_n^2),
\end{equation}
Here $*$ denotes convolution, $\kappa_{\sigma_b}$ is a Gaussian blur kernel, and $\downarrow_s$ denotes bicubic downsampling by factor $s$ followed by bicubic upsampling back to the target size (so that $\mathcal{T}(\mathbf{x})$ stays on the same spatial grid as $\mathbf{x}$).
Therefore, we can get an explicit solution of the target velocity $\mathbf{z}_{1} - \mathbf{x}$ and output velocity $\mathbf{v}_{\theta}\!\left(\mathbf{z}_t,\mathcal{T}(\mathbf{x}),t\right)$ with a single HQ image $\mathbf{x}$ and train the DiT backbone with RFM loss
\begin{equation}
\mathcal{L}_{\mathrm{RFM}}
=\mathbb{E}\Bigl[\bigl\|\mathbf{v}_{\theta}\!\left(\mathbf{z}_t,\mathcal{T}(\mathbf{x}),t\right) - (\mathbf{z}_{1} - \mathbf{x})\bigr\|_2^2\Bigr].
\label{eq:rfm_loss}
\end{equation}
However, for real-world degradation, $\mathcal{T}(\mathbf{x})$ is implicit, which means we can't calculate the loss with a single unpaired HQ image $\mathbf{x}$. So we introduce RCDT in Sec.~\ref{sec:rcdt} to get a numerical approximation of $\mathcal{T}(\mathbf{x})$ again.

\paragraph{Full generation vs.\ latent refinement (optional).}
Our default sampling performs full generation by initializing from pure noise and integrating Eq.~\ref{eq:ode_t} down to $t{=}0$.
For completeness, we also consider a latent refinement variant that initializes at an intermediate time $\beta\in(0,1]$ using the pixel-aligned evidence latent (defined in Sec.~\ref{sec:iiea}):
\begin{equation}
\mathbf{z}_{\beta}=(1-\beta)\,\mathbf{z}^{\mathrm{ctx}}_{\alpha}(\mathbf{y})+\beta\,\boldsymbol{\varepsilon},
\qquad \boldsymbol{\varepsilon}\sim\mathcal{N}(\mathbf{0},\mathbf{I}),
\label{eq:refine_init}
\end{equation}
and then integrates Eq.~\ref{eq:ode_t} from $t{=}\beta$ to $t{=}0$.
$\beta{=}1$ recovers our default full-generation setting, while smaller $\beta$ increases reliance on the evidence initialization.
We analyze the fidelity-realism trade-off of this variant in Sec.~\ref{sec:exp_ablation}.

\subsection{Inpainting-Interface Evidence Adapter (IIEA)}
\label{sec:iiea}

\paragraph{Native FLUX-Fill conditioning.}
FLUX-Fill~\cite{flux1_fill_dev} is trained for inpainting with a pixel-aligned interface.
Given HQ image latent $\mathbf{x} \in\mathbb{R}^{c\times h\times w }$, a zero image (zero in normed [-1, 1] RGB space) latent $\mathbf{x}' \in\mathbb{R}^{c\times h\times w }$, and a binary latent mask $\mathbf{m}\in\{0,1\}^{c' \times H\times W}$ (with $\mathbf{m}{=}1$ indicating the region to be inpainted), the masked-image latent is
\begin{equation}
\mathbf{z}^{\mathrm{masked}}(\mathbf{x},\mathbf{m})
=
(\mathbf{x})\odot(\mathbf{1}-\mathbf{m}) + (\mathbf{x}')\odot\mathbf{m}
\label{eq:zmask_native}
\end{equation}
At noise level $t$, the DiT consumes aligned inputs $(\mathbf{z}_t,\mathbf{z}^{\mathrm{masked}},\mathbf{m})$ on the same spatial grid and concatenated at the channel dimension, which are patchified into token grids without breaking pixel correspondence.

\paragraph{Full-image restoration via evidence injection.}
We convert inpainting into full-image restoration by setting a full-image mask $\mathbf{m}\equiv\mathbf{1}$. This choice is important: in the native inpainting interface, pixels under $\mathbf{m}{=}0$ are treated as preserved context and are therefore encouraged to remain unchanged. Such copy-preserve semantics are mismatched with SR/restoration, where every pixel may need refinement rather than verbatim copying. Under $\mathbf{m}\equiv\mathbf{1}$, the native construction in Eq.~\ref{eq:zmask_native} collapses to an all-zero latent $\mathbf{z}^{\mathrm{masked}}=\mathbf{x}'$ and provides no observation.
IIEA instead \textbf{repurposes} the masked-image slot as a pixel-aligned evidence channel: we keep $\mathbf{m}\equiv\mathbf{1}$ fixed and write the upsampled LQ latent $y$ into $\mathbf{z}^{\mathrm{masked}}$.
Notably, the mask is no longer used as a spatial indicator of known vs.\ unknown pixels; rather, it acts as a constant mode flag that activates the inpainting-conditioning pathway while the dense evidence is carried by the masked-image slot. Although this differs from the pre-training input distribution (masked regions are typically zeroed), LoRA fine-tuning adapts the conditioning pathway to interpret dense evidence.
We optionally modulate the evidence strength with a scalar $\alpha\in[0,1]$ in pixel space:
\begin{equation}
y(\alpha)
  =
\varepsilon\!\left(\alpha \cdot Y\right),
\qquad \alpha\in[0,1],
\label{eq:evidence_strength}
\end{equation}
and define the inpainting-style conditioning bundle
\begin{equation}
\mathbf{c}_{\mathrm{IIEA}}(\mathbf{y};\alpha)
=
\left\{
y(\alpha),\, \mathbf{m}{=}\mathbf{1}
\right\}.
\label{eq:c_iiea}
\end{equation}

\subsection{Lightweight Visual-Semantic Guidance}
\label{sec:redux}

Fill2SR uses Redux~\cite{flux1_redux_dev} as an image embedder instead of a heavyweight VLM. 
Given an input image $\mathbf{y}$, Redux produces a fixed-length semantic token sequence
\begin{equation}
\mathbf{s}(\mathbf{y}) = \mathrm{Redux}(\mathbf{y}),
\end{equation}
which converts images of arbitrary resolution into a 768-token representation. 
These tokens are injected through the model's existing context pathway to provide semantic stabilization while keeping the conditioning cost bounded.

\subsection{Parameter-Efficient Fine-Tuning (LoRA)}
\label{sec:lora}
We freeze the pre-trained DiT backbone and fine-tune only LoRA~\cite{Lora} on attention/FFN projections. For a linear weight $\mathbf{W}$, LoRA adds a rank-$r$ update:
\begin{equation}
\mathbf{W}'=\mathbf{W}+\frac{\lambda}{r}\mathbf{B}\mathbf{A},
\label{eq:lora_def}
\end{equation}
with trainable factors $\mathbf{A},\mathbf{B}$ and scaling $\lambda$ (not the evidence strength $\alpha$ in Sec.~\ref{sec:iiea}). We train $\sim$0.65B parameters; the backbone, VAE, and text encoder are frozen. Target modules and hyperparameters are in the supplementary.

\subsection{RCDT: Reference-Conditioned Degradation Transfer}
\label{sec:rcdt}

Paired HQ/LQ supervision under unknown real-world degradations is scarce, yet our endpoint-flow objective (Eq.~\ref{eq:rfm_loss}) fundamentally relies on paired samples $(\mathbf{x},\mathbf{y} = \mathcal{T}(\mathbf{x}))$ under real-world degradation model $\mathcal{T}$.
RCDT provides such paired endpoint supervision under in-the-wild degradation statistics via reference-anchored degradation transfer, without fitting an explicit parametric degradation model: we use a frozen VLM to distill degradation descriptors from unpaired real-LQ images and a frozen instruction-based image editor~\cite{wu2025qwenimagetechnicalreport} to transfer these degradations onto clean targets, synthesizing paired observations $\hat{\mathbf{y}}\approx\mathcal{T}(\mathbf{x})$.
Given an unpaired real-LQ set $\{\mathbf{y}^{\mathrm{real}}_i\}_{i=1}^{M}$, we mine one degradation description per image using Qwen3-VL~\cite{Qwen3-VL} to form a prompt pool $\mathcal{P}$.
For each clean training image $\mathbf{x}$ (used as the ground-truth target), we sample $\mathbf{d}\!\sim\!\mathrm{Unif}(\mathcal{P})$ and let Qwen3-VL produce a content-aware specification $\tilde{\mathbf{d}}$ describing how $\mathbf{d}$ would manifest on $\mathbf{x}$.
Finally, an instruction-based editor~\cite{wu2025qwenimagetechnicalreport} transfers the specified degradations onto $\mathbf{x}$ to synthesize the paired LQ observation $\hat{\mathbf{y}}$:
\begin{equation}
\mathbf{d}_i=\mathrm{Qwen3VL}(\mathbf{y}^{\mathrm{real}}_i),\;
\tilde{\mathbf{d}}=\mathrm{Qwen3VL}(\mathbf{x},\mathbf{d}),\;
\hat{\mathbf{y}}=\mathrm{Edit}(\mathbf{x};\tilde{\mathbf{d}}).
\label{eq:rcdt_pipeline}
\end{equation}
RCDT is an offline data-synthesis step with frozen models (no test-time overhead); we include its compute in end-to-end cost. To isolate architectural gains from extra supervision, we report results \emph{with/without} RCDT (Tabs.~\ref{tab:real_sr_scales},~\ref{tab:ablation_main}). As RCDT aims to transfer real-world degradation statistics, it primarily improves in-the-wild perceptual quality and may trade off distortion-oriented full-reference metrics on synthetic benchmarks. We therefore adopt a conservative construction and usage protocol:
(i) degradation-only prompting (no new objects/text and no intentional geometry edits),
(ii) VLM-based verification to reject content drift or severe artifacts,
(iii) leakage control by keeping the real-LQ pool disjoint from evaluation sets and screening near-duplicates,
and (iv) \textbf{late-stage injection}: we mix RCDT pairs only in the final 20\% of training to refine toward real-world statistics while preserving the synthetic fidelity learned earlier.
More details are provided in the supplementary material.

\section{Experiments}
\label{sec:experiments}

\subsection{Experimental Setup}
\label{sec:exp_setup}
 
\textbf{Datasets and Metrics.}
We evaluate on synthetic (DIV2K~\cite{DIV2K}, LSDIR~\cite{LSDIR}, FFHQ~\cite{FFHQ}) and real-world benchmarks (RealLQ250~\cite{DreamClear}, RealPhoto60~\cite{SUPIR, FaithDiff}).
Following FaithDiff~\cite{FaithDiff}, we construct three synthetic degradation levels (D1/D2/D3) of increasing severity by combining downsampling with stronger blur, additive noise, and JPEG compression, corresponding to $2\times$, $4\times$, and $8\times$ inputs.
On synthetic benchmarks we report PSNR$\uparrow$, SSIM$\uparrow$, LPIPS$\downarrow$, and DINO feature similarity to audit semantic consistency (higher is better).
On real benchmarks without ground truth, we report MUSIQ$\uparrow$~\cite{MUSIQ}, MANIQA$\uparrow$~\cite{MANIQA}, and CLIPIQA$\uparrow$~\cite{CLIPIQA}; we additionally report AES on RealLQ250 (score and induced rank).

\noindent\textbf{Baselines and Evaluation Protocol.}
We compare against Real-ESRGAN~\cite{wang2021realesrgan}, StableSR~\cite{StableSR}, DiffBIR~\cite{DiffBIR}, OSEDiff~\cite{wu2024one}, SeeSR~\cite{SeeSR}, SUPIR~\cite{SUPIR}, DreamClear~\cite{DreamClear}, FaithDiff~\cite{FaithDiff}, and DiT4SR~\cite{dit4sr}.

\noindent\textbf{Backbone capacity note.} These methods rely on backbones of different scales/modalities (e.g., SD2/SDXL, PixArt, and SD3.5). We evaluate each pipeline using its official checkpoint and recommended settings, and report memory/compute footprints (Tab.~\ref{tab:efficiency_32g} and Supplementary) to contextualize capacity differences.
For diffusion/flow baselines, we use their default sampling and (if available) default tiling configuration; for prompt-conditioned methods, we use their default automatic prompt generation without manual editing.
For real-world SR, we further enforce a unified output-resolution budget by evaluating at long-side targets of 512/1024/2048 pixels, corresponding to $2\times/4\times/8\times$ on RealLQ250 and $2\times/4\times$ on RealPhoto60 (higher-resolution inputs).
\subsection{Implementation Details}
\label{sec:exp_impl}

\textbf{Backbone and Objective.}
Fill2SR is built upon FLUX-Fill~\cite{flux1_fill_dev,Flux}.
We perform restoration in the latent space of a frozen VAE and learn a conditional rectified flow using the RFM objective in Eq.~\ref{eq:rfm_loss}.
Low-quality evidence is injected through our proposed inpainting-interface evidence adapter (Sec.~\ref{sec:iiea}), and Redux semantics (Sec.~\ref{sec:redux}) are enabled by default.
We freeze the DiT backbone and train only LoRA parameters~\cite{Lora} (rank $256$) on attention and FFN projections.
The total number of trainable parameters is 0.65B.

\noindent\textbf{Training Schedule and Data Mixture.}
We train for 20k steps with a learning rate of $2.5{\times}10^{-4}$, a 2.5k-step warm-up, and a global batch size of 16. Rather than adapting at a single fixed resolution, we use mixed-resolution, mixed-aspect-ratio crops: $512$-$1536$ pixels in the early stage and $1024$-$2048$ pixels in the later stage, with 1:1, 4:3, and 16:9 buckets up to native QHD ($2560\times1440$). We pre-train on 300k synthetic pairs and inject 10k RCDT-generated pairs only in the final stage. This recipe, enabled by the bounded conditioning cost of IIEA, is a key reason for the stable behavior across 512/1024/2048 test scales. RCDT data synthesis is conducted offline and introduces no test-time overhead.

\noindent\textbf{Compute and Memory Optimizations.}
Model training requires $4\times$ RTX 5880 Ada (48GB) for $14$ days ($50.6$ A100e GPU-days; $56$ raw GPU-days; see Supplementary for normalization).
RCDT data generation uses $4\times$ RTX 5880 Ada for $5$ days ($18.1$ A100e GPU-days; $20$ raw GPU-days).
To enable native QHD training within 48GB VRAM, we precompute and cache all frozen features (VAE latents, text-encoder embeddings when used, and Redux tokens), and enable gradient checkpointing in the DiT backbone.

\subsection{Main Results}
\label{sec:exp_results}

\noindent\textbf{Quantitative comparison on synthetic benchmarks.}
Table~\ref{tab:synth_summary} reports results averaged over three degradation levels (D1--D3) on DIV2K~\cite{DIV2K}, LSDIR~\cite{LSDIR}, and FFHQ~\cite{FFHQ}.
As expected, distortion-oriented fidelity metrics (PSNR/SSIM) tend to favor regression-based pipelines.
To isolate the effect of domain-transfer paired supervision, we report \textit{Ours-Base} (w/o RCDT) and \textit{Ours-Full} (+RCDT).
\textit{Ours-Base} achieves the best LPIPS on DIV2K and LSDIR and remains competitive on FFHQ-face.
After injecting RCDT pairs, \textit{Ours-Full} trades a small amount of synthetic-domain LPIPS for consistently stronger perceptual/no-reference metrics (MUSIQ/MANIQA/CLIPIQA), indicating improved recovery of photo-realistic textures on real degradations.

\noindent\textbf{Quantitative comparison on real-world SR and the effect of RCDT.}
Table~\ref{tab:real_sr_scales} reports results on RealLQ250~\cite{DreamClear} and RealPhoto60~\cite{SUPIR} under multiple upscaling factors, and includes DiT4SR~\cite{dit4sr} as a strong diffusion baseline.
Consistent with the synthetic-to-real trade-off observed in Table~\ref{tab:synth_summary}, injecting RCDT incurs only a slight drop in synthetic-domain full-reference metrics (e.g., LPIPS) but yields substantial improvements on real-world perceptual/no-reference metrics.
While DiT4SR attains strong results at the extreme $8\times$ setting on RealLQ250, \textit{Ours-Full} achieves the best or second-best performance on most entries across scales.
These results support the fidelity-preserving control of IIEA and the effectiveness of RCDT for real-world domain transfer.

\noindent\textbf{Aesthetic quality and semantic consistency audit.}
No-reference IQA metrics can sometimes reward over-sharpening or hallucinated details. To better assess whether perceptual gains come with semantic drift, we additionally report (i) AES on RealLQ250 (rank/score) and (ii) DINO feature consistency on paired benchmarks.
Table~\ref{tab:story_main_d3_summary} summarizes the AES score and ranking, and reports DINO consistency and robustness on LSDIR\_VAL. Fill2SR achieves the best RealLQ250 AES score and maintains the second best DINO similarity.

\begin{figure}[h]
    \centering
    \includegraphics[width=\textwidth]{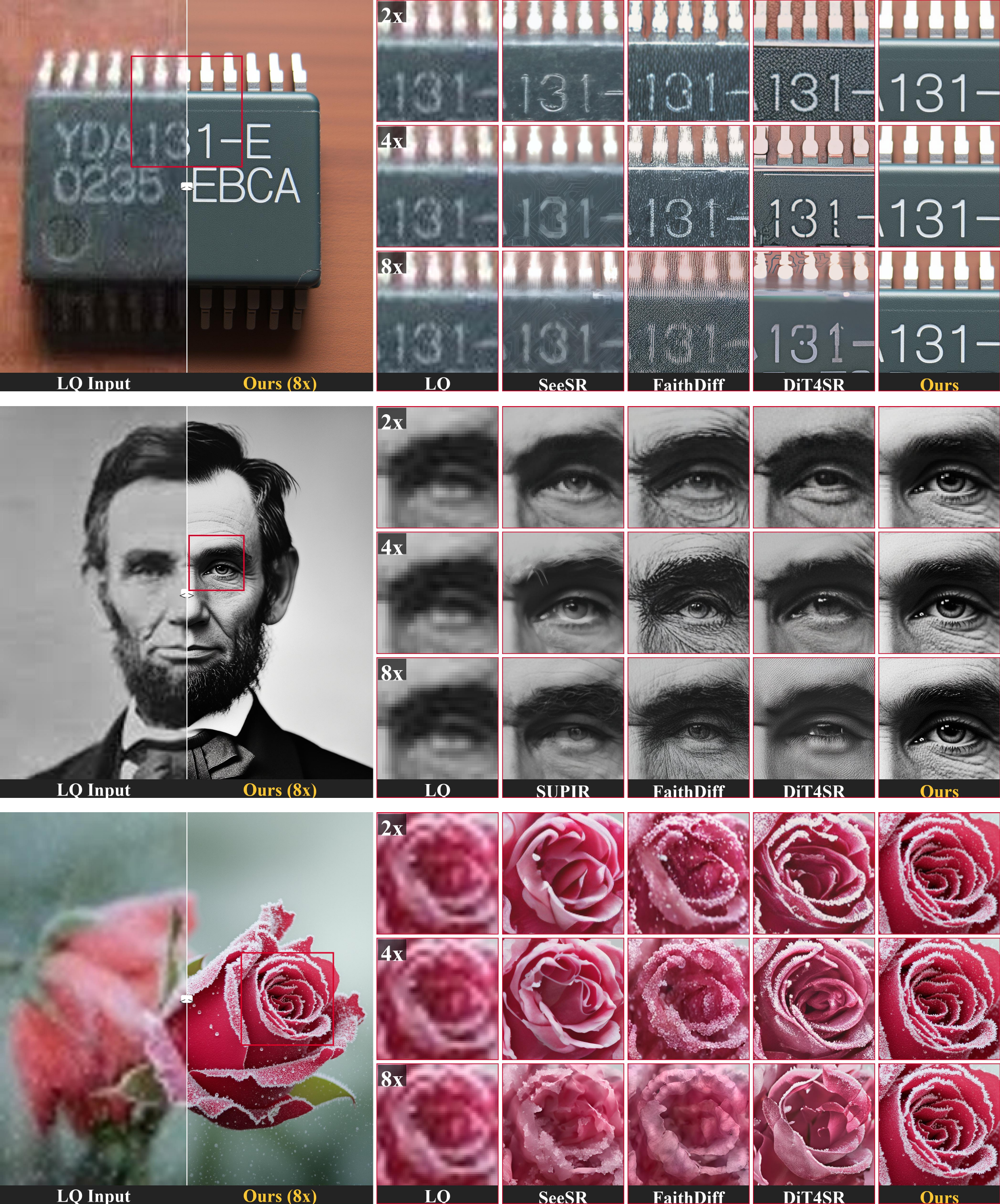}
    \caption{
        \textbf{Qualitative comparisons on real-world scenarios at up to $8\mathbf{x}$ degradation.} 
        Fill2SR produces visually plausible restorations across diverse and challenging domains.
        \textbf{Row 1 (Industrial Text):} Fill2SR can improve text legibility and boundary sharpness, while some baselines introduce distracting structured artifacts.
        \textbf{Row 2 (Natural Textures):} Fill2SR tends to recover fine textures (e.g., frost-like micro-structures) without excessive over-smoothing.
        \textbf{Row 3 (Human Faces):} Fill2SR often preserves facial structure and eye highlights, while some baselines exhibit local geometric drift.
        The left split-screen intuitively highlights our global visual enhancement.
        More results with additional baselines are provided in the Supplementary Material.
    }
    \label{fig:qual_realworld}
\end{figure}

\noindent\textbf{Qualitative comparisons.}
Fig.~\ref{fig:qual_realworld} shows representative results on real-world inputs.
Fill2SR improves text legibility and recovers natural micro-textures (e.g., frost crystals) without compromising structural details (e.g., facial geometry).
Due to space, Fig.~\ref{fig:qual_realworld} focuses on representative diffusion/flow-based baselines (including DiT4SR~\cite{dit4sr}) and highlights robustness across scale factors (2×/4×/8×).
Additional qualitative comparisons (including synthetic D3 cases and a broader set of baselines) are provided in the Supplementary Material.

%

\begin{table}[t]
  \centering
\caption{
  \textbf{Quantitative comparison on synthetic benchmarks (\textit{DIV2K}, \textit{LSDIR}, and \textit{FFHQ}).}
  Results are averaged across three degradation levels (D1--D3).
  While regression-based methods obtain higher PSNR/SSIM, our generative approach performs strongly on perceptual metrics (e.g., LPIPS, CLIPIQA), indicating improved recovery of photo-realistic textures.
  \textbf{Ours-Base} is trained with standard synthetic paired data only (w/o RCDT), whereas \textbf{Ours-Full} additionally injects RCDT supervision (+RCDT) in the final stage, which may trade off synthetic-domain LPIPS while boosting perceptual/no-reference scores.
  Blue outlines mark \textbf{Ours-Full} on LPIPS for readability.
  The best and second-best results are highlighted in \textbf{bold} and \underline{underlined}.
}
  \label{tab:synth_summary}
  \renewcommand{\arraystretch}{1.0}
  \setlength{\tabcolsep}{3.2pt}
  \small
  \begin{adjustbox}{max width=0.9\textwidth}
  \begin{tabular}{l l *{9}{S[table-format=2.4]}}
        \toprule
        \multirow{3}{*}{Datasets} & \multirow{3}{*}{Metrics} & \multicolumn{9}{c}{Methods} \\
        \cmidrule(lr){3-11}
        &  & \mch{Real-ESRGAN} & \mch{StableSR} & \mch{DiffBIR} & \mch{SeeSR} & \mch{SUPIR} & \mch{DreamClear} & \mch{FaithDiff} & \mch{\makecell{\textbf{Ours-Base}\\\scriptsize (w/o RCDT)}} & \mch{\makecell{\textbf{Ours-Full}\\\scriptsize (+RCDT)}} \\
        \midrule

      \multirow{6}{*}{\textit{DIV2K\_VAL}}
        & PSNR$\uparrow$& \secbest{23.3520} & 22.9703 & \best{23.4959} & 22.5903 & 22.6610 & 21.7065 & 22.4013 & \synthcell{22.4596} & 21.7831 \\
        & SSIM$\uparrow$& \best{0.6414} & \secbest{0.6085} & 0.6065 & 0.5807 & 0.5748 & 0.5557 & 0.5574 & \synthcell{0.5872} & 0.5679 \\
        & \textbf{LPIPS}$\downarrow$& \lpipscell{0.4017} & \lpipscell{0.4380} & \lpipscell{0.4168} & \lpipscell{0.4165} & \lpipscell{0.4032} & \lpipscell{0.4153} & \lpipscell{0.4051} & \synthcellDark{\best{0.3866}} & \lpipscellbluebox{\secbest{0.3981}} \\
        & MUSIQ$\uparrow$& 58.9699 & 49.0826 & 66.1573 & \best{69.0464} & 64.6142 & 65.9251 & \secbest{68.1122} & 65.9644 & \fullcell{67.8893} \\
        & MANIQA$\uparrow$& 0.5252 & 0.4444 & 0.5709 & 0.6015 & 0.5858 & 0.5743 & \secbest{0.6184} & 0.6173 & \fullcell{\best{0.6271}} \\
        & CLIPIQA$\uparrow$& 0.4875 & 0.3801 & 0.5737 & \secbest{0.6035} & 0.5051 & 0.5310 & 0.5763 & 0.6011 & \fullcell{\best{0.6138}} \\
      \midrule

      \multirow{6}{*}{\textit{LSDIR\_VAL}}
        & PSNR$\uparrow$& \best{20.7295} & 20.3530 & \secbest{20.7116} & 20.0585 & 19.9397 & 19.4056 & 19.6345 & \synthcell{19.6930} & 19.2354 \\
        & SSIM$\uparrow$& \best{0.5670} & 0.5185 & \secbest{0.5287} & 0.4900 & 0.4847 & 0.4923 & 0.4622 & \synthcell{0.5019} & 0.4890 \\
        & \textbf{LPIPS}$\downarrow$& \lpipscell{0.4055} & \lpipscell{0.4442} & \lpipscell{0.4090} & \lpipscell{0.4126} & \lpipscell{0.4202} & \lpipscell{0.4100} & \lpipscell{0.4083} & \synthcellDark{\best{0.3878}} & \lpipscellbluebox{\secbest{0.3974}} \\
        & MUSIQ$\uparrow$& 64.5601 & 52.8770 & 70.1140 & \secbest{72.6831} & 67.9508 & 70.3541 & 71.6307 & 72.3820 & \fullcell{\best{73.1206}} \\
        & MANIQA$\uparrow$& 0.5674 & 0.4788 & 0.6153 & 0.6388 & 0.6234 & 0.6149 & 0.6653 & \secbest{0.6840} & \fullcell{\best{0.6859}} \\
        & CLIPIQA$\uparrow$& 0.5424 & 0.4199 & 0.6309 & 0.6413 & 0.5704 & 0.6121 & 0.6307 & \secbest{0.7153} & \fullcell{\best{0.7223}} \\
      \midrule

      \multirow{6}{*}{\textit{FFHQ-face}}
        & PSNR$\uparrow$& \best{28.3966} & 27.3251 & \secbest{27.9839} & 27.3894 & 27.4845 & 26.6292 & 26.4585 & \synthcell{26.5658} & 26.3979 \\
        & SSIM$\uparrow$& \best{0.7777} & 0.7190 & 0.7291 & \secbest{0.7305} & 0.7040 & 0.6902 & 0.6689 & \synthcell{0.7081} & 0.7097 \\
        & \textbf{LPIPS}$\downarrow$& \lpipscell{0.3838} & \lpipscell{0.3879} & \lpipscell{0.4016} & \lpipscell{\secbest{0.3731}} & \lpipscell{\best{0.3682}} & \lpipscell{0.3837} & \lpipscell{0.3768} & \synthcellDark{0.3827} & \lpipscell{0.3857} \\
        & MUSIQ$\uparrow$& 63.4199 & 66.7731 & 73.9354 & 73.3842 & 73.6046 & 70.7586 & \secbest{76.1941} & 75.6041 & \fullcell{\best{76.4407}} \\
        & MANIQA$\uparrow$& 0.4920 & 0.5029 & 0.5925 & 0.5875 & 0.5970 & 0.5654 & \best{0.6406} & 0.6219 & \fullcell{\secbest{0.6286}} \\
        & CLIPIQA$\uparrow$& 0.4297 & 0.4827 & 0.6283 & 0.5493 & 0.5300 & 0.4807 & 0.5719 & \secbest{0.6355} & \fullcell{\best{0.6470}} \\
      \bottomrule
    \end{tabular}
  \end{adjustbox}
\end{table}

\begin{table}[t]
  \centering
\caption{
  \textbf{Quantitative comparison on real-world SR benchmarks (\mbox{\textit{RealPhoto60}} and \mbox{\textit{RealLQ250}}).}
  We upscale real-world inputs without explicit degradation priors and evaluate at target output sizes of 512/1024/2048 pixels (\mbox{\textit{RealLQ250}}: $2\times/4\times/8\times$; \mbox{\textit{RealPhoto60}}: $2\times/4\times$).
  \textbf{Ours-Full} (+RCDT; 10k pairs injected in the final stage) consistently improves over \textbf{Ours-Base} (w/o RCDT) on perceptual/no-reference metrics.
  The best and second-best results are highlighted in \textbf{bold} and \underline{underlined}.
}
  \label{tab:real_sr_scales}
  \renewcommand{\arraystretch}{1.0}
  \setlength{\tabcolsep}{2.8pt}
  \small
  \begin{adjustbox}{max width=\textwidth}
\begin{tabular}{l c l *{10}{S[table-format=2.4]} >{\columncolor{eccvblue!8}}S[table-format=2.4]}
      \toprule
      \multirow{2}{*}{Datasets} & \multirow{2}{*}{Scale} & \multirow{2}{*}{Metrics} & \multicolumn{11}{c}{Methods} \\
      \cmidrule(lr){4-14}
      & & &
      \mch{Real-ESRGAN} & \mch{StableSR} & \mch{DiffBIR} & \mch{OSEDiff} & \mch{SeeSR} & \mch{SUPIR} & \mch{DreamClear} & \mch{FaithDiff} & \mch{DiT4SR} & \mch{\makecell{\textbf{Ours-Base}\\\scriptsize (w/o RCDT)}} & \mch{\makecell{\textbf{Ours-Full}\\\scriptsize (+RCDT)}} \\
      \midrule

      \multirow{6}{*}{\textit{RealPhoto60}}
        & \multirow{3}{*}{$2\times$}
          & MUSIQ$\uparrow$ & 59.0296 & 50.2670 & 61.6646 & 70.4686 & 71.8052 & 69.6326 & 68.8353 & 71.5853 & \secbest{72.5731} & 70.6389 & \best{73.5949} \\
        & & MANIQA$\uparrow$& 0.4797 & 0.4554 & 0.5617 & 0.5891 & 0.6079 & 0.6116 & 0.5905 & \best{0.6513} & 0.6309 & 0.6307 & \secbest{0.6468} \\
        & & CLIPIQA$\uparrow$& 0.5068 & 0.3632 & 0.5465 & 0.5726 & 0.5959 & 0.5972 & 0.5383 & 0.5718 & 0.6192 & \secbest{0.6780} & \best{0.7137} \\
      \addlinespace[2pt]
        & \multirow{3}{*}{$4\times$}
          & MUSIQ$\uparrow$ & 50.9528 & 22.4571 & 49.8760 & 55.4632 & 58.1401 & 57.0462 & 62.4641 & 58.5360 & \secbest{63.7202} & 61.8281 & \best{65.8413} \\
        & & MANIQA$\uparrow$& 0.4785 & 0.2960 & 0.5522 & 0.5321 & 0.5530 & 0.5269 & 0.5077 & 0.5657 & 0.5803 & \secbest{0.5957} & \best{0.6042} \\
        & & CLIPIQA$\uparrow$& 0.4247 & 0.2354 & 0.4739 & 0.4595 & 0.5386 & 0.4542 & 0.3765 & 0.5007 & \best{0.5879} & 0.5207 & \secbest{0.5486} \\
      \midrule
      \multirow{9}{*}{\textit{RealLQ250}}
        & \multirow{3}{*}{$2\times$}
          & MUSIQ$\uparrow$ & 67.7327 & 63.0388 & 71.5459 & 71.3742 & \secbest{72.1775} & 68.3565 & 69.4476 & 71.6023 & 71.7326 & 67.5721 & \best{72.5417} \\
        & & MANIQA$\uparrow$& 0.6177 & 0.6116 & 0.6676 & 0.6567 & 0.6643 & 0.6463 & 0.6499 & \best{0.6869} & 0.6760 & 0.6454 & \secbest{0.6861} \\
        & & CLIPIQA$\uparrow$& 0.5784 & 0.4904 & \secbest{0.6435} & 0.6116 & 0.6363 & 0.6196 & 0.6356 & 0.6299 & 0.6244 & 0.6373 & \best{0.7158} \\
      \addlinespace[2pt]
        & \multirow{3}{*}{$4\times$}
          & MUSIQ$\uparrow$ & 62.5154 & 50.4880 & 65.9055 & 69.5570 & 70.3728 & 63.2153 & 66.2208 & 69.8047 & \secbest{70.5121} & 62.1914 & \best{71.5498} \\
        & & MANIQA$\uparrow$& 0.5239 & 0.4500 & 0.5700 & 0.5782 & 0.5927 & 0.5812 & 0.5756 & \secbest{0.6373} & 0.6154 & 0.5751 & \best{0.6387} \\
        & & CLIPIQA$\uparrow$& 0.4359 & 0.3332 & 0.5370 & 0.5282 & 0.5568 & 0.4542 & 0.5017 & 0.5269 & \secbest{0.5617} & 0.5384 & \best{0.6287} \\
      \addlinespace[2pt]
        & \multirow{3}{*}{$8\times$}
          & MUSIQ$\uparrow$ & 41.9557 & 25.0598 & 50.9256 & 57.0877 & 58.9324 & 51.3011 & 49.7753 & 58.1177 & \best{61.8378} & 50.0516 & \secbest{59.3031} \\
        & & MANIQA$\uparrow$& 0.4582 & 0.3119 & 0.5274 & 0.5196 & 0.5509 & 0.5173 & 0.4991 & 0.5653 & \secbest{0.5766} & 0.5516 & \best{0.5876} \\
        & & CLIPIQA$\uparrow$& 0.3614 & 0.2425 & 0.4425 & 0.4435 & \secbest{0.5071} & 0.3849 & 0.3881 & 0.4829 & \best{0.5403} & 0.4293 & 0.4784 \\
      \bottomrule
    \end{tabular}
  \end{adjustbox}
\end{table}

\begin{table*}[t]
  \centering
  \begin{adjustbox}{width=0.8\textwidth}
\begin{tabular}{l ccccccc >{\columncolor{eccvblue!8}}S[table-format=2.4]}
      \toprule
      \mch{Metric} & \mch{StableSR} & \mch{DiffBIR} & \mch{SUPIR} & \mch{DreamClear} & \mch{SeeSR} & \mch{FaithDiff} & \mch{DiT4SR} & \mch{\textbf{Ours-Full}} \\
      \midrule
      RealLQ250 Score$\uparrow$ & 4.9950 & 5.2653 & 5.2718 & 5.3254 & 5.3296 & 5.2487 & \secbest{5.5432} & \best{5.5510} \\
      RealLQ250 Rank$\downarrow$ & 7.9160 & 5.9640 & 5.5760 & 5.0080 & 4.6800 & 5.7280 & \best{2.7560} & \secbest{2.9400} \\
      LSDIR DINO Avg$\uparrow$ & 0.6667 & 0.7452 & 0.7397 & \best{0.8089} & 0.7676 & 0.7694 & 0.7604 & \secbest{0.8018} \\
      \bottomrule
    \end{tabular}
  \end{adjustbox}
  \caption{Additional results on aesthetic quality and semantic consistency. RealLQ250 is evaluated on the 4x results. DINO Avg is computed by averaging DINO over LSDIR\_VAL from D1 to D3. Best and second-best are marked with \textbf{bold} and \underline{underlined}.}
  \label{tab:story_main_d3_summary}
\end{table*}

\begin{table}[b]
  \centering
  \small
  \setlength{\tabcolsep}{4pt}
  \begin{adjustbox}{max width=0.8\textwidth}
  \begin{tabular}{lccccc}
    \toprule
    Method & Params (Train/Backbone, B) & Time (s)$\downarrow$ & 32GB & Train (A100e)$\downarrow$ & Data (A100e)$\downarrow$ \\
    \midrule
    SeeSR (SD2)        & 2.30 / 2.30 & 34  & \cmark & N/R & N/A \\
    SUPIR (SDXL)       & 1.30 / 1.30 & 26  & \xmark & 317.5 (640) & N/A \\
    DreamClear (PixArt)& 2.41 / 2.41 & 107 & \xmark & 224.0 (224) & 592.8 (1360) \\
    FaithDiff (SDXL)   & 2.40 / 2.40 & 35  & \xmark & N/R & N/A \\
    \rowcolor{eccvblue!8}
    \textbf{Fill2SR (FLUX-Fill)} & \textbf{0.65 / 12.0} & 75 & \textbf{\cmark} & \textbf{50.6 (56)} & \textbf{18.1 (20)} \\
    \bottomrule
  \end{tabular}
\end{adjustbox}
  \caption{
  Cost breakdown and 32GB feasibility at $1536^2$.
  Time is measured with 28 inference steps (batch=1) under each method's recommended inference configuration (including its default tiling strategy and prompt pipeline when applicable).
  32GB indicates whether the full pipeline runs without out-of-memory errors on a single 32GB GPU at $1536^2$.
  ``Train'' and ``Data'' report A100-equivalent GPU-days (A100e) as the primary value; raw GPU-days are shown in parentheses. A100e uses peak dense BF16/FP16 tensor throughput for coarse cross-hardware normalization (see Supplementary for details). N/R and N/A denote not reported and not applicable.
  }
  \label{tab:efficiency_32g}
\end{table}

\paragraph{Memory scalability and adaptation cost.}
As shown in Table~\ref{tab:efficiency_32g}, Fill2SR fits within a single 32GB GPU at $1536^2$ while keeping the conditioning budget constant.
Although we rely on a large frozen backbone and external models for offline RCDT synthesis, the end-to-end compute budget (training + data generation) remains substantially lower than prior diffusion restoration pipelines under the reported settings.
For a more meaningful cross-hardware comparison, Table~\ref{tab:efficiency_32g} reports A100-equivalent GPU-days (A100e) as the main value and raw GPU-days in parentheses; the Supplementary Material details the normalization.

\FloatBarrier 
\subsection{Ablation Study}
\label{sec:exp_ablation}

\begin{figure}[t]
  \centering
  \IfFileExists{plot/results/ablation_grid.pdf}{
    \includegraphics[width=0.8\textwidth]{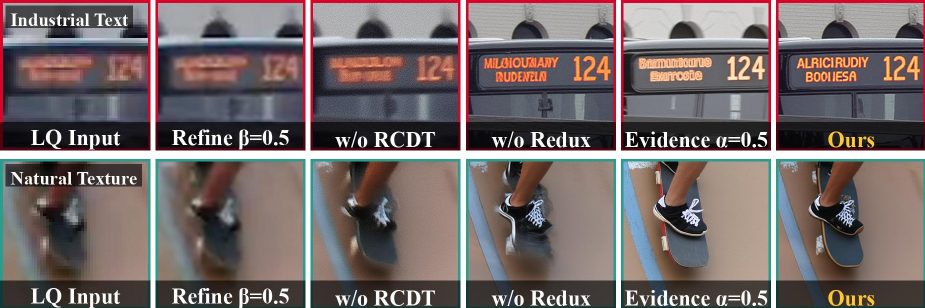}
  }{
    \fbox{\parbox[c][0.18\textheight][c]{0.96\textwidth}{
      \centering
      \textbf{Ablation qualitative figure placeholder.}\\
      Put \texttt{plot/results/ablation\_qualitative.pdf} here.
    }}
  }
  \caption{
    \textbf{Qualitative ablation on representative real-world inputs (RealLQ250).}
    Columns show the LQ input, latent refinement (\mbox{$\beta=0.5$}), w/o RCDT, w/o Redux, weakened evidence injection (\mbox{$\alpha=0.5$}), and the default Fill2SR output.
    \emph{Top:} industrial text. \emph{Bottom:} natural texture.
  }
  \label{fig:ablation_qual}
\end{figure}

We ablate the key components of Fill2SR on synthetic (DIV2K~\cite{DIV2K}, LSDIR~\cite{LSDIR}) and real-world (RealLQ250~\cite{DreamClear}) sets, averaging scores over three degradation levels (D1--D3). Detailed per-level results are in the Supplementary Material.

\noindent\textbf{Evidence strength $\alpha$ (IIEA).} $\alpha$ controls how strongly the model uses pixel-aligned evidence without changing the conditioning budget. Table~\ref{tab:ablation_main} shows that smaller $\alpha$ degrades both synthetic distortion and real-image no-reference scores. Differences between $\alpha\in\{0.9,1.0\}$ are minor, making it a mild robustness knob rather than a dramatic controller. We default to $\alpha{=}1$.

\noindent\textbf{Redux visual-semantic tokens.} Removing Redux tokens degrades no-reference metrics under severe degradations, indicating that bounded semantic guidance prevents category drift and stabilizes global structure. Alternative choices (zero/noise image tokens) underperform Redux, supporting our use of the image embedder and removal of the VLM.

\noindent\textbf{RCDT paired synthesis.} RCDT provides reference-anchored paired endpoint supervision complementing standard synthetic degradations. Removing it significantly degrades real-world no-reference metrics on RealLQ250, while slightly improving synthetic distortion metrics (\eg, DIV2K\_VAL PSNR/LPIPS). This indicates RCDT primarily aligns the training supervision to real degradation statistics rather than ``gaming'' synthetic metrics; we further report semantic/aesthetic audits to probe metric bias and hallucinations (Tab.~\ref{tab:story_main_d3_summary}). RCDT is offline only and introduces zero test-time overhead.

\noindent\textbf{Full generation vs.\ latent refinement.} Partial refinement ($\beta{<}1$) improves synthetic PSNR/SSIM but severely harms real-image perceptual quality. We hypothesize intermediate initializations fall off the learned flow manifold, conflicting with pixel-aligned evidence and yielding unstable trajectories~\cite{SDEdit}. Thus, we use full generation ($\beta{=}1$) by default.

\begin{table}[t]
  \centering
  \small
  \setlength{\tabcolsep}{4pt}
  \renewcommand{\arraystretch}{1.15}
  \begin{adjustbox}{max width=0.8\textwidth}
    \begin{tabular}{lccccccc}
      \toprule
      Setting & \multicolumn{2}{c}{DIV2K\_VAL} & \multicolumn{2}{c}{LSDIR\_VAL} & \multicolumn{3}{c}{RealLQ250} \\
      \cmidrule(lr){2-3} \cmidrule(lr){4-5} \cmidrule(lr){6-8}
      Setting & PSNR$\uparrow$ & LPIPS$\downarrow$ & PSNR$\uparrow$ & LPIPS$\downarrow$ & CLIPIQA$\uparrow$ & MANIQA$\uparrow$ & MUSIQ$\uparrow$ \\
      \midrule
      \rowcolor{eccvblue!8}
      \textbf{Default} & 21.7831 & \secbest{0.3981} & 19.2354 & \secbest{0.3974} & \best{0.6076} & \best{0.6375} & \best{67.7982} \\
      w/o RCDT data & 22.4596 & \best{0.3866} & 19.6930 & \best{0.3878} & 0.5350 & 0.5907 & 59.9384 \\
      \midrule
      w/o Redux tokens & 21.1379 & 0.4125 & 18.6203 & 0.4118 & 0.5963 & 0.6328 & 67.1587 \\
      Redux tokens (zero image) & 21.4694 & 0.4067 & 18.9790 & 0.4081 & 0.6023 & 0.6336 & 67.4110 \\
      Redux tokens (noise image) & 21.6229 & 0.4047 & 19.0723 & 0.4043 & \secbest{0.6059} & 0.6355 & 67.4399 \\
      \midrule
      Evidence $\alpha{=}0.5$ & 20.9871 & 0.4105 & 18.5959 & 0.4040 & 0.6031 & 0.6333 & 67.5953 \\
      Evidence $\alpha{=}0.7$ & 21.4020 & 0.4042 & 18.9017 & 0.4004 & 0.6037 & 0.6348 & 67.5593 \\
      Evidence $\alpha{=}0.9$ & 21.7308 & 0.3994 & 19.1563 & 0.3981 & 0.6058 & \secbest{0.6364} & \secbest{67.6596} \\
      \midrule
      Refine $\beta{=}0.5$ & 23.7536 & 0.5568 & 21.1851 & 0.5731 & 0.3047 & 0.3987 & 28.2379 \\
      Refine $\beta{=}0.7$ & \best{23.8974} & 0.5261 & \best{21.2508} & 0.5501 & 0.3075 & 0.3972 & 28.8641 \\
      Refine $\beta{=}0.9$ & \secbest{23.8655} & 0.4433 & \secbest{21.2463} & 0.4599 & 0.3314 & 0.4027 & 34.7317 \\
      \bottomrule
    \end{tabular}
  \end{adjustbox}
  \caption{Ablation results on DIV2K\_VAL, LSDIR\_VAL, and RealLQ250, averaged over three degradation levels (D1--D3). We compare removing RCDT training data, controlling Redux semantic tokens (disabled / zero-image / noise-image), and sweeping evidence/refine strengths.}
  \label{tab:ablation_main}
\end{table}

\subsection{Limitations and Discussion}
\label{sec:limitations}
Fill2SR is a generative restorer and may synthesize plausible yet incorrect details when the evidence is extremely weak or ambiguous; it should not be used in scenarios requiring pixel-level authenticity (\eg, forensics, medical imaging, or legal evidence).
Although the adaptation is parameter-efficient in \emph{trainable} parameters, inference still requires running a large frozen backbone (\eg, 12B for FLUX-Fill-dev~\cite{flux1_fill_dev}) and iterative sampling, which is slower than feed-forward regressors-distillation and faster solvers are important directions.
IIEA repurposes the masked-image slot in a way that differs from the backbone's original inpainting pre-training; while LoRA fine-tuning mitigates this mismatch, a deeper analysis of the resulting conditioning behavior is left for future work.
RCDT relies on frozen instruction-based editors and VLM verification, which may still miss subtle geometric drift; we adopt conservative prompting and filtering, but cannot fully guarantee pixel-perfect alignment for every synthesized pair.
Finally, because sampling is stochastic, a fuller multi-seed statistical characterization of no-reference metrics would further strengthen the evaluation.

\section{Conclusion}
We presented Fill2SR as a complementary inpainting-interface formulation for SR that keeps conditioning overhead bounded with resolution while making memory use predictable, without relying on ControlNet-style branches or spatial token streams.
IIEA repurposes the native masked-image interface of inpainting diffusion transformers and writes VAE-encoded LQ evidence into this slot, enabling mixed-resolution training up to native QHD ($2560\times1440$). With reverse-degradation rectified-flow training, bounded Redux guidance, and offline RCDT synthesis, Fill2SR achieves strong LPIPS performance on synthetic benchmarks and strong perceptual quality on real-world benchmarks while remaining feasible on a single 32GB GPU at $1536^2$. Code and models will be released at \url{https://github.com/Xingfu-Yi/Fill2SR}.
\nocite{hosu2020koniq10k,fang2020spaq,LAION5B}
%
%
\FloatBarrier 

%% file: appendix.tex
\section{Additional Visual Results}

\begin{center}
  \centering
  \includegraphics[width=\textwidth]{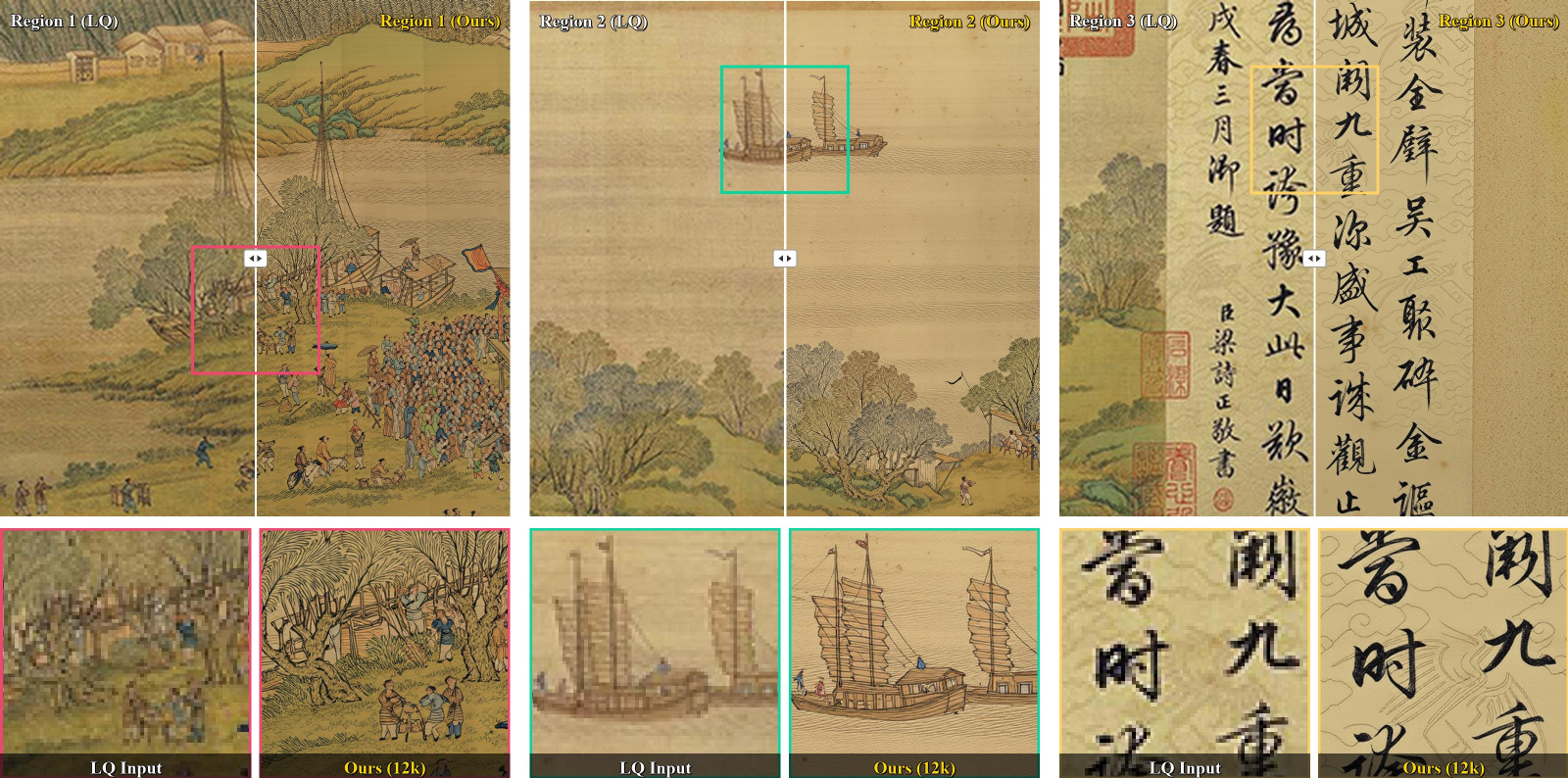}
  \captionof{figure}{\textbf{Ultra-high-resolution (12k) real-world restoration showcase.}
Three regions from \textit{Along the River During the Qingming Festival} are shown as LQ/Fill2SR split views (top), with the corresponding marked-region crops below. The examples highlight object structure, textures, and calligraphic strokes. This result uses tiled inference and does not imply native 12k training.}
  \label{fig:12k_demo}
\end{center}

\clearpage

\begin{center}
  \centering
  \includegraphics[width=\textwidth]{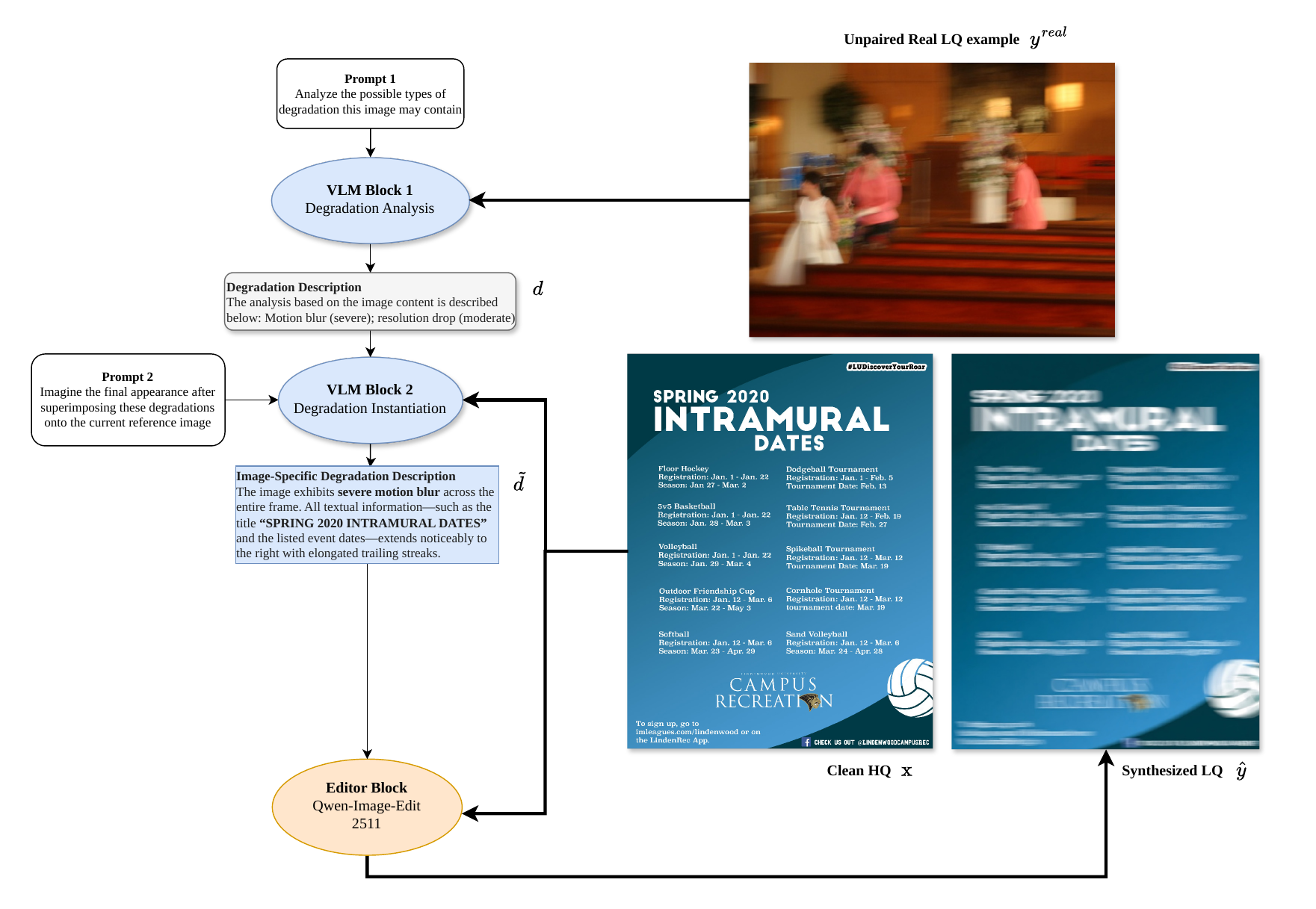}
  \captionof{figure}{\textbf{Illustration of the RCDT paired data synthesis pipeline.}
  Given an unpaired real-world LQ image $\mathbf{y}^{\mathrm{real}}$, a frozen VLM first analyzes its visible degradations and produces a natural-language degradation description $d$.
  Conditioned on $d$ and a clean HQ reference image $\mathbf{x}$, the VLM then refines $d$ into an image-specific degraded-appearance instruction $\tilde d$ for $\mathbf{x}$.
  A frozen instruction-based image editor applies $\tilde d$ to $\mathbf{x}$ to synthesize a paired LQ image $\hat{\mathbf{y}}$, while keeping $\mathbf{x}$ as the HQ supervision target.
  For readability, the prompts and VLM outputs shown here are abbreviated and lightly paraphrased to convey the core intent of the actual pipeline.}
  \label{fig:rcdt_pipeline}
\end{center}

\clearpage

\begin{center}
    \centering
    \includegraphics[width=0.85\textwidth]{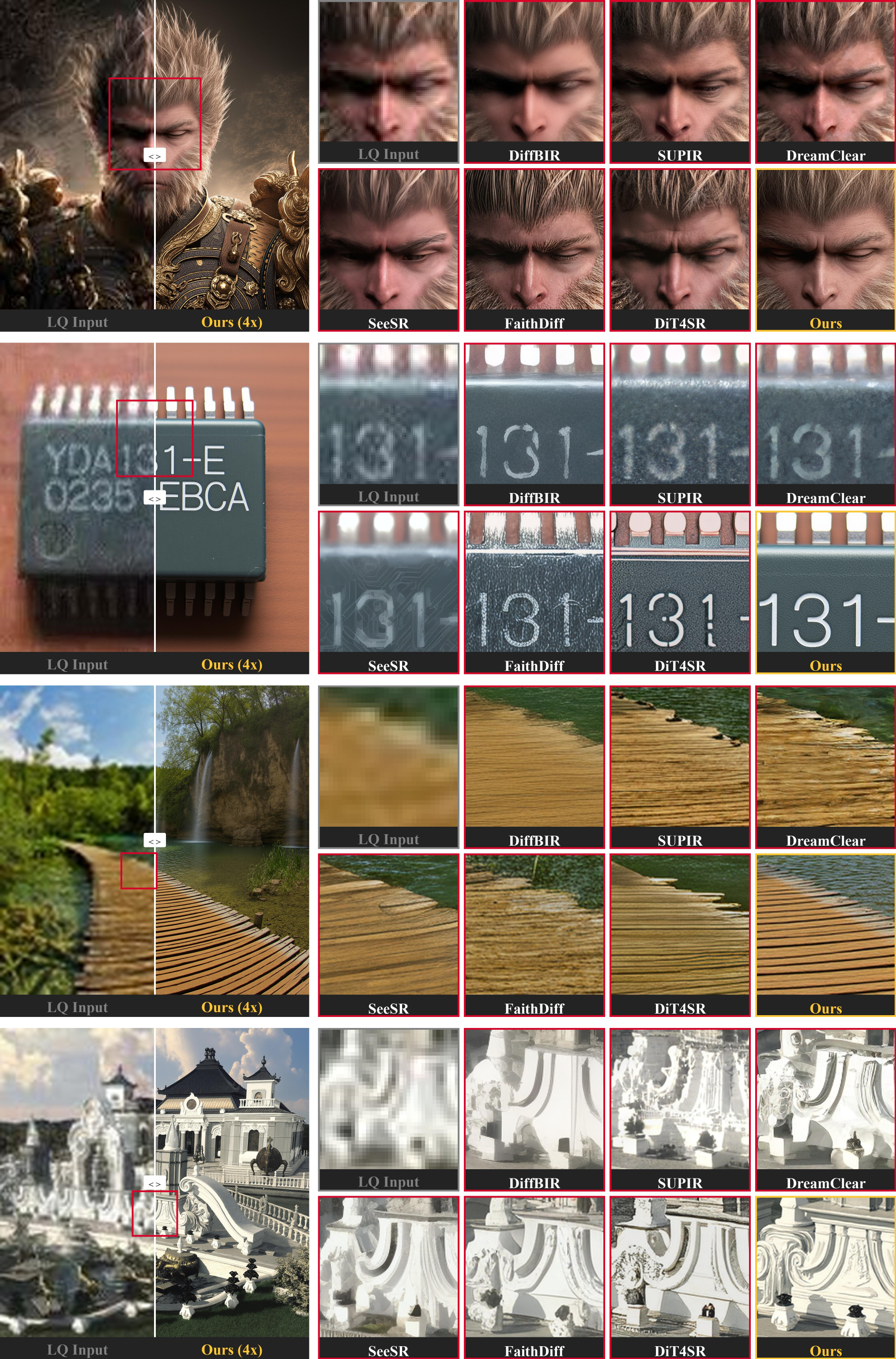}
    \captionof{figure}{\textbf{Additional qualitative comparisons on RealLQ250 ($4\times$).}
    Rows show a portrait, electronic component, wooden bridge, and architectural detail. Fill2SR shows finer skin/hair textures, a cleaner component surface, clearer planks, and sharper marble patterns.
    Left: LQ/Fill2SR overview and marked crop region. Right: LQ, DiffBIR, SUPIR, DreamClear, SeeSR, FaithDiff, DiT4SR, and Ours (Fill2SR). No paired ground truth is available.}
  \label{fig:supp_real_grid_1}
\end{center}

\clearpage

\begin{center}
    \centering
    \includegraphics[width=0.85\textwidth]{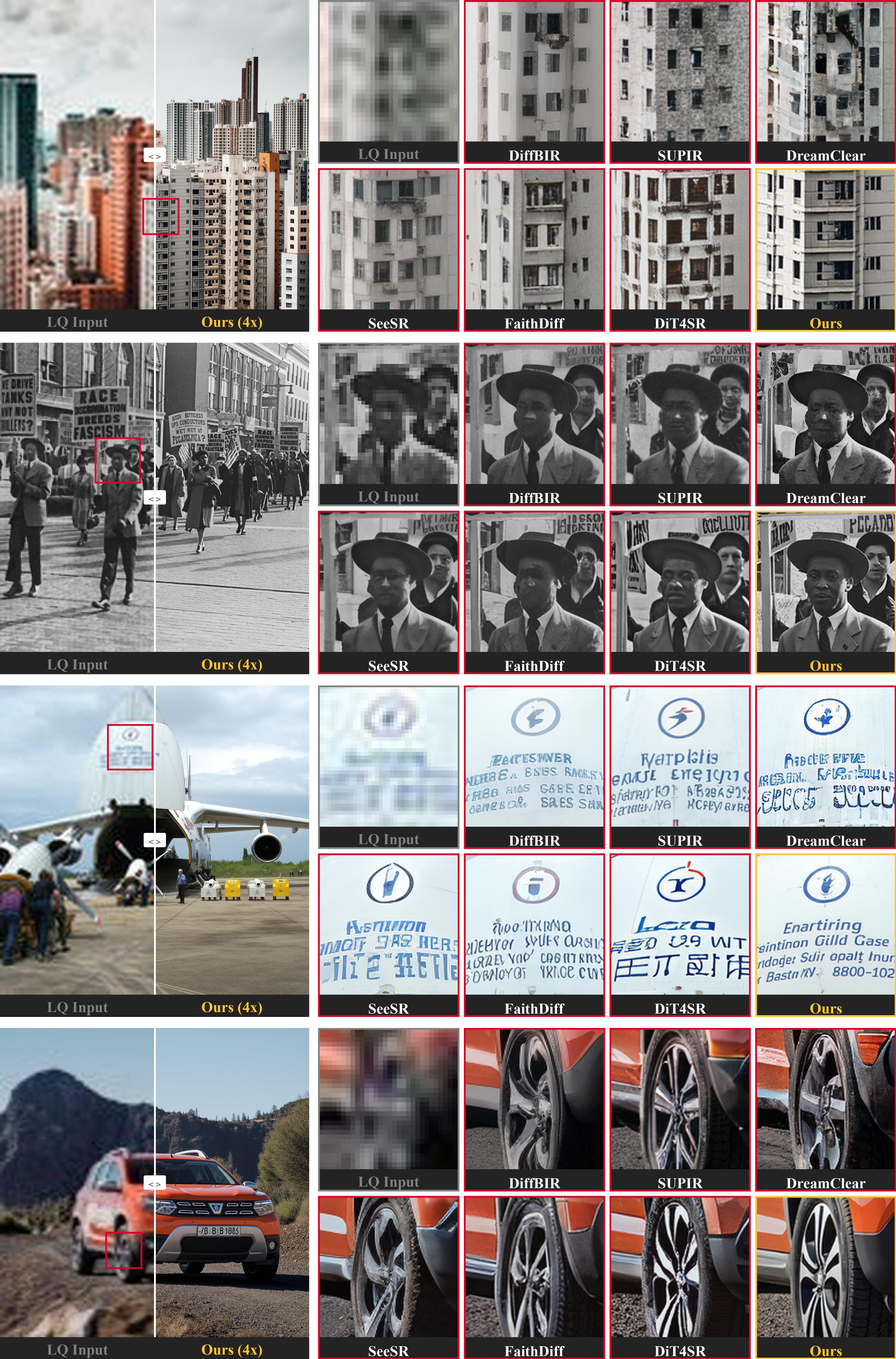}
    \captionof{figure}{\textbf{Additional qualitative comparisons on RealLQ250 ($4\times$).}
    Rows show building details, a face, aircraft lettering, and a wheel rim. The Fill2SR crops show fewer spurious building textures, clearer facial details and lettering, and more coherent rim geometry.
    Left: LQ/Fill2SR overview and marked crop region. Right: LQ and the labeled baseline/Fill2SR crops, using the same methods as Fig.~\ref{fig:supp_real_grid_1}. No paired ground truth is available.}
  \label{fig:supp_real_grid_2}
\end{center}

\clearpage

\begin{center}
    \centering
    \includegraphics[width=0.85\textwidth]{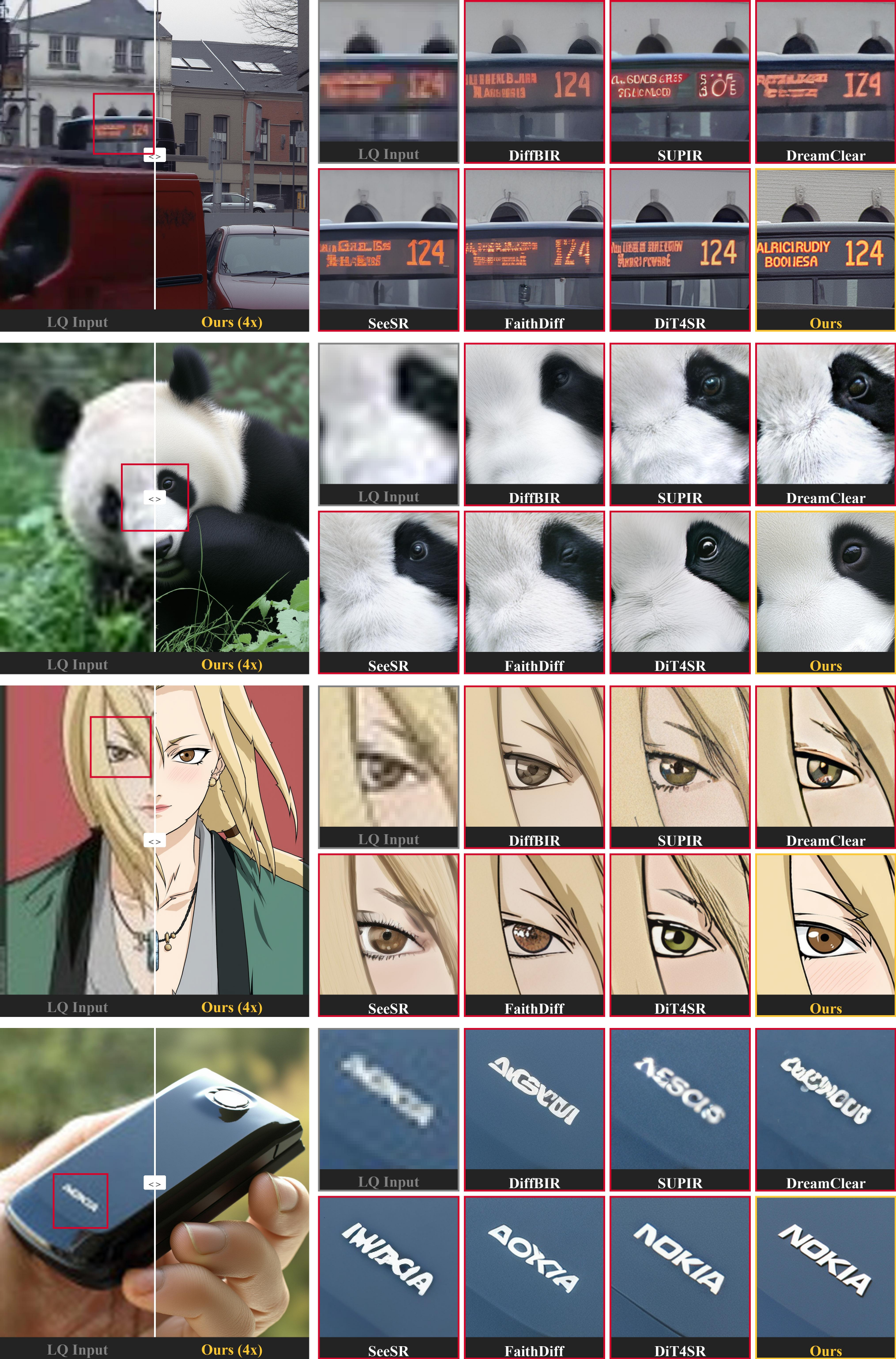}
    \captionof{figure}{\textbf{Additional qualitative comparisons on RealLQ250 ($4\times$).}
    Rows show bus-route text, a panda, an anime portrait, and phone lettering. Fill2SR shows more legible text, clearer eye boundaries, and cleaner line work; the phone crop contains legible \textit{NOKIA} lettering.
    Left: LQ/Fill2SR overview and marked crop region. Right: the labeled method crops, using the same methods as Fig.~\ref{fig:supp_real_grid_1}. No paired ground truth is available.}
  \label{fig:supp_real_grid_4}
\end{center}

\clearpage

\begin{center}
    \centering
    \includegraphics[width=0.85\textwidth]{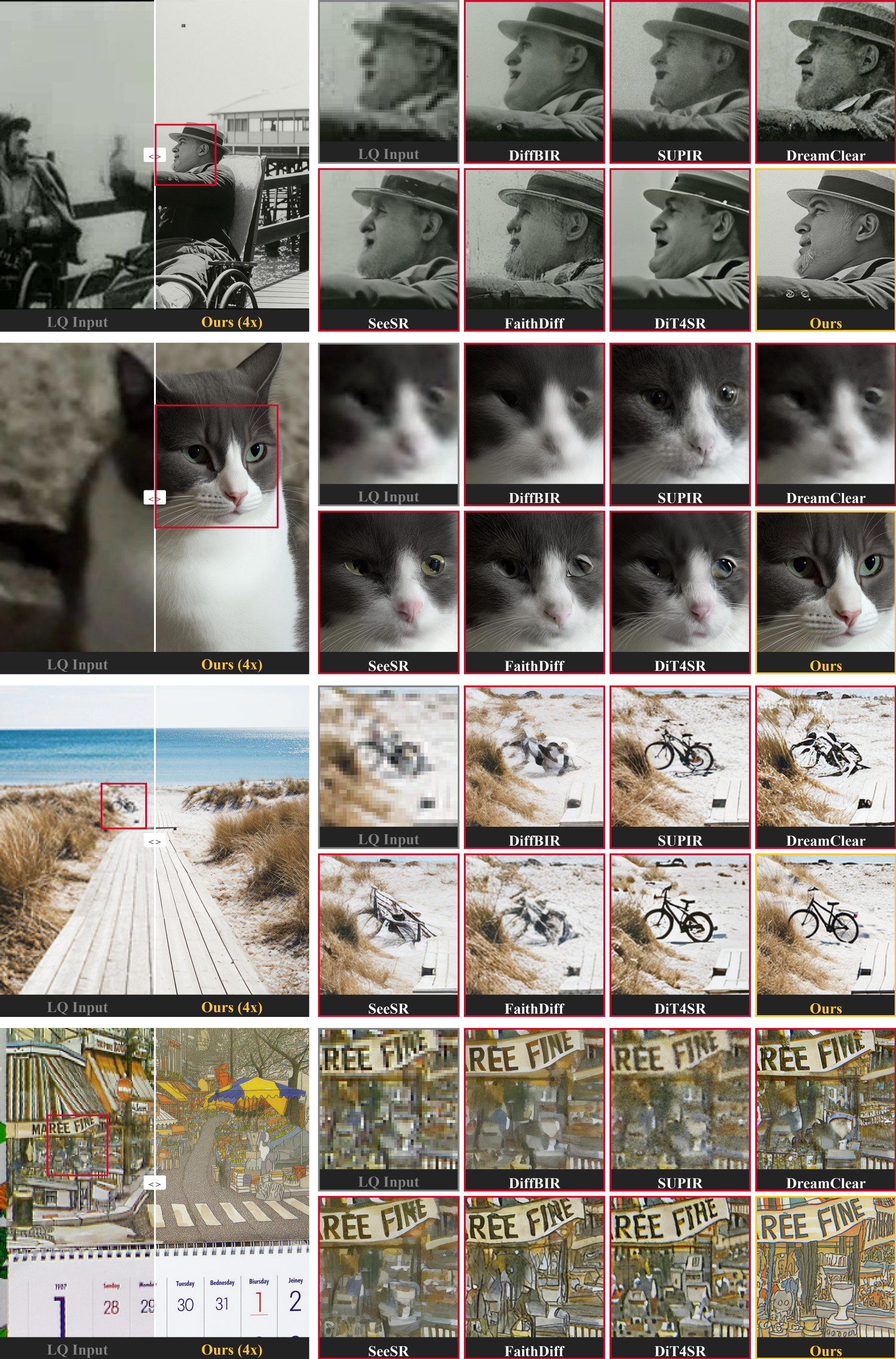}
    \captionof{figure}{\textbf{Additional qualitative comparisons on RealLQ250 ($4\times$).}
    Rows show an old portrait, a cat, a bicycle, and a stylized illustration. The Fill2SR crops show cleaner face/eye regions, more coherent bicycle geometry, and sharper lines and lettering, while several baseline crops contain blur or structured artifacts.
    Left: LQ/Fill2SR overview and marked crop region. Right: the labeled method crops, using the same methods as Fig.~\ref{fig:supp_real_grid_1}. No paired ground truth is available.}
  \label{fig:supp_real_grid_5}
\end{center}

\clearpage

\begin{center}
    \centering
    \includegraphics[width=0.85\textwidth]{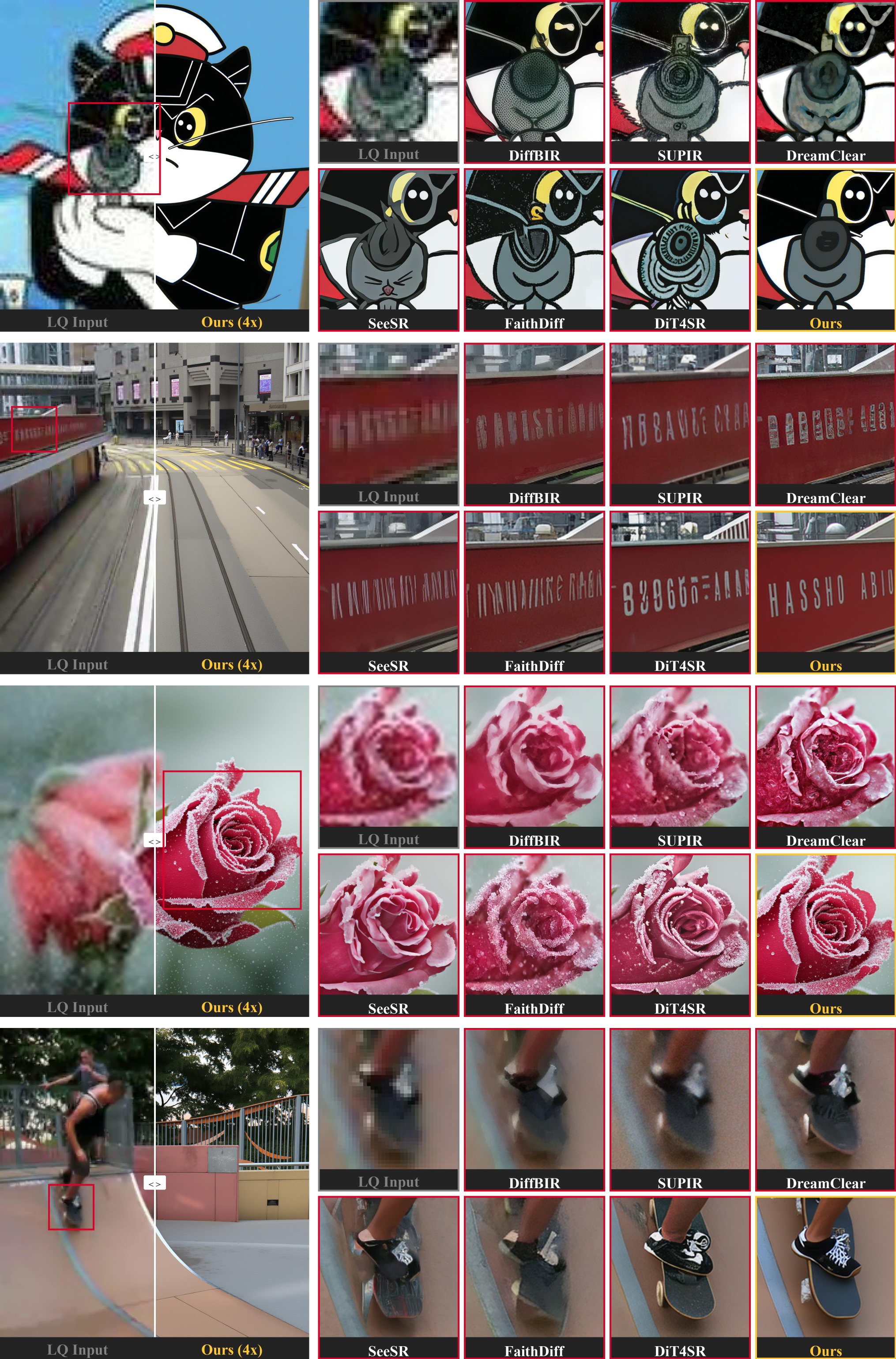}
\captionof{figure}{\textbf{Additional qualitative comparisons on RealLQ250 ($4\times$).}
Rows show a cartoon, a banner, a rose, and a skateboard scene. Fill2SR shows sharper lines and banner text, clearer separation of frost-like texture from petals, and more coherent shoe/foot/skateboard structure.
Left: LQ/Fill2SR overview and marked crop region. Right: the labeled method crops, using the same methods as Fig.~\ref{fig:supp_real_grid_1}. No paired ground truth is available.}
  \label{fig:supp_real_grid_6}
\end{center}

\clearpage

\begin{center}
    \centering
    \includegraphics[width=\textwidth]{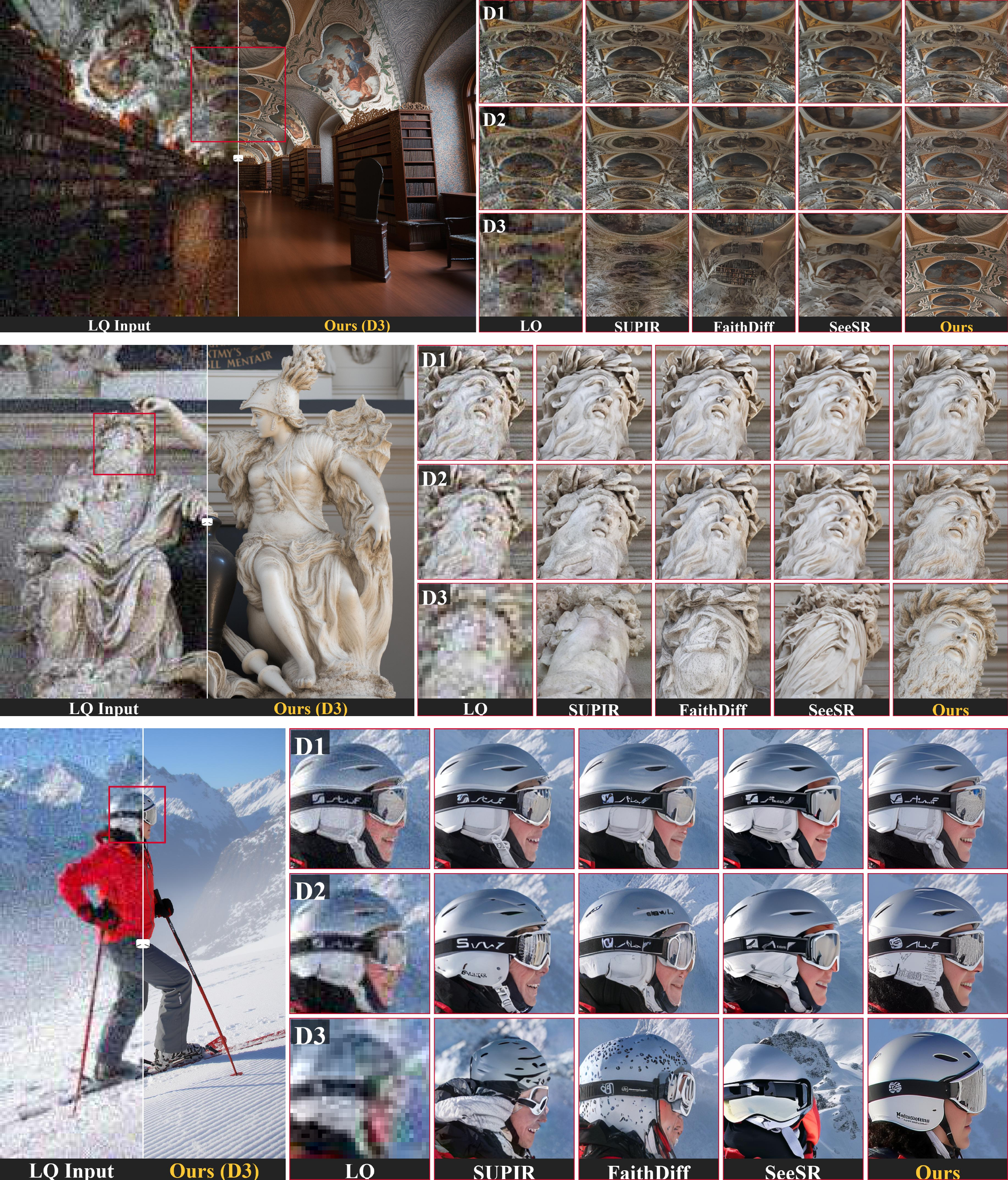}
\captionof{figure}{\textbf{Additional synthetic comparisons across D1--D3 ($2\times/4\times/8\times$).}
Left: split views of the D3 LQ input and Fill2SR output, with marked crop regions. Right: D1--D3 crops for LQ, SUPIR, FaithDiff, SeeSR, and Ours (Fill2SR).
Under stronger degradation, several baselines introduce ceiling patterns, lose statue-face structure, or distort helmet/face details. The displayed Fill2SR crops retain clearer mural structure and more stable facial and helmet geometry.}
    \label{fig:supp_qual_d3}
\end{center}